%% file: main.tex
\documentclass[fleqn,10pt]{wlscirep}
\usepackage[utf8]{inputenc}
\usepackage[T1]{fontenc}
\usepackage{booktabs, tabularx, ragged2e, enumitem}

\begin{document}

\title{Efficient Deep Learning for Medical Imaging: Bridging the Gap Between High-Performance AI and Clinical Deployment}

\author[1]{Cuong Manh Nguyen}
\author[1,*]{Truong-Son Hy}

\affil[1]{Department of Computer Science, University of Alabama at Birmingham, Alabama, United States}

\affil[*]{Corresponding author. \href{thy@uab.edu}{thy@uab.edu}}

\input{tex/abstract}

\flushbottom
\maketitle
\input{tex/introduction}
\input{tex/related_work}
\input{tex/taxonomy}
\input{tex/optimization}
\input{tex/discussion}
\input{tex/conclusion}
\thispagestyle{empty}

\bibliography{main}

\section*{Author Contributions}

T.S.H. conceived the study and supervised the project. C.M.N. conducted the literature review and wrote the majority of the manuscript. T.S.H. provided guidance on the scope and structure of the review and edited the manuscript. All authors reviewed and approved the final manuscript.

\section*{Funding}

This research received no specific grant from any funding agency in the public,
commercial, or not-for-profit sectors.

\section*{Competing Interests}

The authors declare no competing interests.

\end{document}

%% file: tex/abstract.tex
\begin{abstract}
Deep learning has revolutionized medical image analysis, playing a vital role in modern clinical applications. However, the deployment of large-scale models in real-world clinical settings remains challenging due to high computational costs, latency constraints, and patient data privacy concerns associated with cloud-based processing. To address these bottlenecks, this review provides a comprehensive synthesis of efficient and lightweight deep learning architectures specifically tailored for the medical domain. We categorize the landscape of modern efficient models into three primary streams: Convolutional Neural Networks (CNNs), Lightweight Transformers, and emerging Linear Complexity Models. Furthermore, we examine key model compression strategies (including pruning, quantization, knowledge distillation, and low-rank factorization) and evaluate their efficacy in maintaining diagnostic performance while reducing hardware requirements. By identifying current limitations and discussing the transition toward on-device intelligence, this review serves as a roadmap for researchers and practitioners aiming to bridge the gap between high-performance AI and resource-constrained clinical environments.  
\end{abstract}

%% file: tex/introduction.tex
\section{Introduction} \label{sec:introduction}

The convergence of Artificial Intelligence (AI) and healthcare has fundamentally reshaped the landscape of modern medicine, driven by the availability of high-dimensional medical imaging modalities and increasingly sophisticated learning algorithms. From radiology and pathology to ophthalmology and cardiology, deep learning systems have demonstrated expert-level performance across a wide range of diagnostic tasks, often surpassing human benchmarks under controlled experimental conditions~\cite{topol2019high, esteva2019guide}. These advances have fueled optimism regarding the transformative potential of AI in improving diagnostic accuracy, streamlining clinical workflows, and alleviating workforce shortages.

However, the dominant paradigm underpinning many of these successes has been the relentless scaling of model size, depth, and computational complexity. Inspired by breakthroughs in natural image analysis and large-scale foundation models, medical imaging research has increasingly embraced architectures with hundreds of millions, or even billions of parameters. While such scaling often yields incremental gains in benchmark performance, it comes at a rapidly escalating computational cost. This trend has reached a critical inflection point, where marginal improvements in accuracy are outweighed by substantial increases in training time, memory footprint, inference latency, and energy consumption.

The operational implications of this imbalance are profound. Large-scale medical imaging models demand specialized hardware accelerators, continuous maintenance, and significant energy resources, leading to prohibitive deployment costs for many healthcare systems. Beyond economic considerations, these models raise growing concerns regarding environmental sustainability. Recent analyses indicate that the carbon footprint associated with training and deploying state-of-the-art deep learning models can be substantial, with emissions comparable to those generated by multiple automobiles over their lifetime~\cite{selvan2022carbon, truhn2024ecological}. Such findings have catalyzed the emergence of the ``Green AI'' movement, which advocates for efficiency-aware model development as a core research objective rather than an afterthought~\cite{henderson2020towards}. In the context of healthcare, where ethical responsibility extends beyond individual patients to population-level impact, this shift is particularly salient.

Crucially, high algorithmic performance does not automatically translate into clinical utility. Despite impressive results in retrospective studies, the real-world integration of AI systems into clinical workflows remains limited. This gap reflects a fundamental performance paradox (the deployment paradox), wherein models optimized for accuracy under idealized conditions fail to meet the practical constraints of clinical environments. These constraints can be conceptualized as a persistent “trilemma” encompassing data privacy, latency requirements, and infrastructure inequality.

First, stringent privacy regulations and ethical considerations severely restrict cloud-based processing of sensitive medical data. Contrary to long-standing assumptions, medical images such as MRI scans are not inherently anonymous. Strikingly, recent studies demonstrate that facial reconstruction and recognition techniques can re-identify MRI participants with accuracies reaching up to $83\%$, exposing patients to significant re-identification risks~\cite{schwarz2019identification, giouroukou2025rethinking}. As a result, transmitting raw imaging data to centralized cloud servers is increasingly viewed as incompatible with modern data protection frameworks, including the General Data Protection Regulation (GDPR) and the Health Insurance Portability and Accountability Act of 1996 (HIPAA). These concerns necessitate a shift toward on-device or edge-based inference, where data remain within institutional or point-of-care boundaries~\cite{kaissis2020secure}.

Second, latency constraints impose strict real-time requirements in many clinical scenarios. In applications such as robot-assisted surgery, interventional imaging, and Point-of-Care Ultrasound (POCUS), inference delays of even a few hundred milliseconds can degrade clinical performance or compromise patient safety~\cite{anvari2005impact, choksi2023bringing}. While cloud offloading is computationally attractive, it cannot reliably guarantee low-latency responses due to network congestion, bandwidth variability, and connectivity disruptions. This makes lightweight, locally executable models a prerequisite rather than a design preference.

Third, infrastructure inequality poses a major barrier to the equitable deployment of AI-driven diagnostics. Many rural hospitals and healthcare facilities in low-income and middle-income countries lack access to high-end GPUs, stable electricity, or high-bandwidth internet connections. As a result, computationally intensive models exacerbate existing disparities in global health by disproportionately benefiting well-resourced institutions~\cite{guo2018application}. Addressing this imbalance requires AI systems that are robust, efficient, and deployable across heterogeneous hardware platforms, including low-power edge devices~\cite{rajput2024ai}.

\begin{figure*}[ht]
\centering
\includegraphics[width=1.0\textwidth]{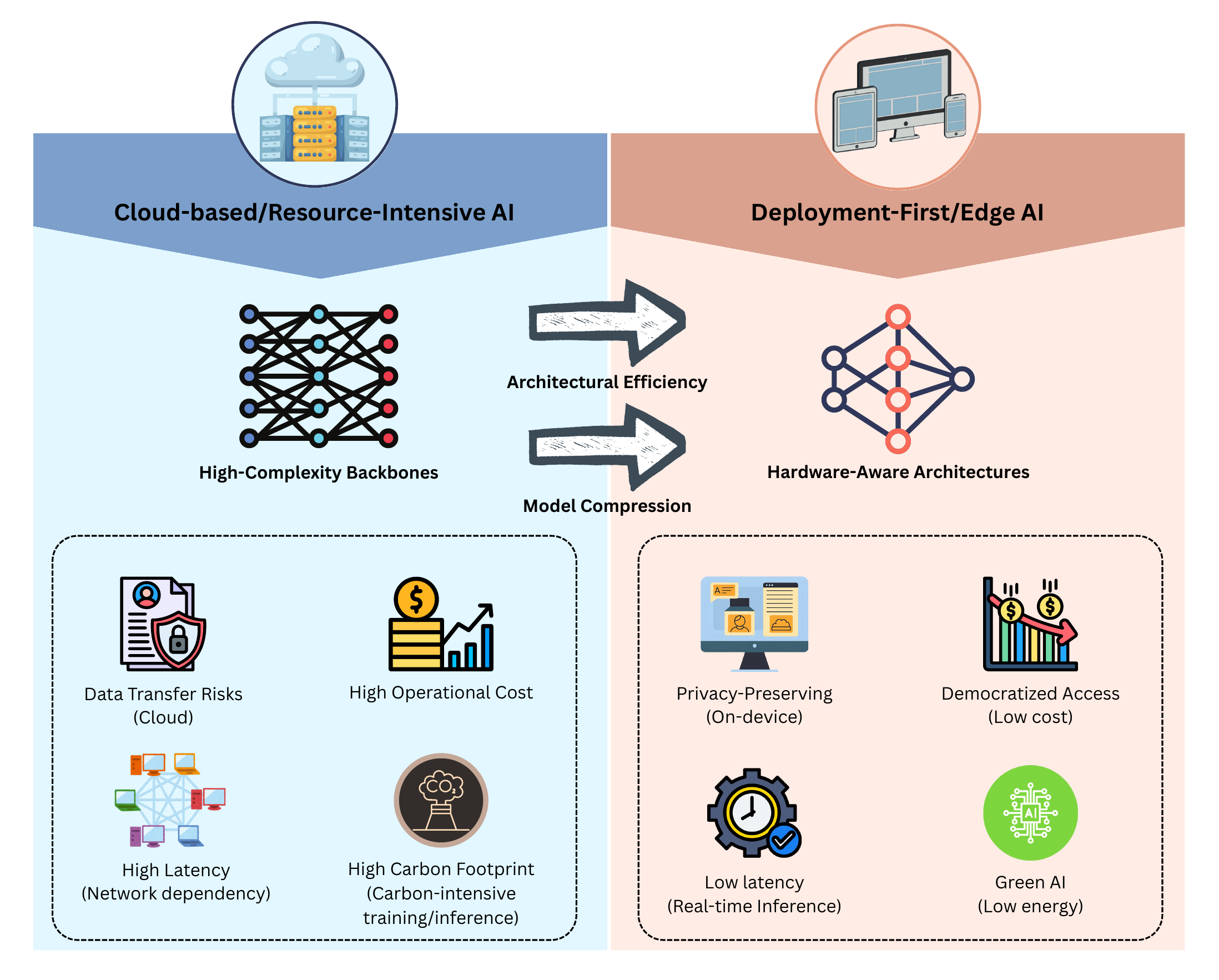}
\caption{The Paradigm Shift toward Deployment-First Medical AI. The left boundary illustrates the limitations of centralized, resource-intensive models, including privacy risks and high latency. The right boundary showcases the improvements achieved through efficient architectures and model compression, enabling secure, real-time, and sustainable clinical deployment.}
\label{fig:main}
\end{figure*}

Collectively, these challenges necessitate a fundamental paradigm shift in medical AI research. As illustrated in Figure \ref{fig:main}, the field is moving away from a reliance on resource-intensive, cloud-centric models toward a ``deployment-first'' framework. This perspective contrasts the limitations of the traditional approach, such as privacy risks and latency bottlenecks, with the improvements offered by efficient, edge-native designs. Consequently, efficiency is reframed not merely as a secondary optimization step, but as a central criterion that determines whether AI systems can bridge the gap between algorithmic potential and real-world clinical value.

Motivated by this perspective, this review synthesizes the rapidly evolving landscape of efficient deep learning for medical imaging. We provide a structured roadmap that spans both architectural innovation and model optimization strategies aimed at bridging the gap between algorithmic performance and clinical deployment. Specifically, we categorize architectural advances into three major streams: (1) efficient convolutional neural networks (CNNs), which leverage inductive biases and parameter sharing for resource-aware learning; (2) lightweight Transformer-based models that seek to mitigate the quadratic complexity of self-attention; and (3) emerging linear-complexity architectures, including Mamba and state space models, which promise scalable global context modeling for high-resolution medical images. Complementing this architectural analysis, we examine key model compression techniques, such as network pruning, quantization, knowledge distillation, and low-rank factorization. These techniques further reduce computational overhead while preserving diagnostic fidelity within clinically acceptable margins. By critically analyzing the trade-offs among model size, energy consumption, latency, and clinical sensitivity, this review aims to guide researchers and practitioners toward the development of AI systems that are not only high-performing, but also truly deployable in resource-constrained clinical environments.

%% file: tex/related_work.tex
\section{Related Work}
\label{sec:related-work}

The rapid adoption of deep learning in medical imaging has led to a substantial body of survey literature. Existing reviews provide valuable perspectives on architectural innovation and learning paradigms, yet they are predominantly organized around specific model families or algorithmic advances. As a result, questions concerning deployment readiness under clinical resource constraints are often addressed only implicitly. This section situates the present survey within this evolving literature.

\subsection{Architecture-Centric Surveys in Medical Imaging}

A large fraction of existing reviews focuses on individual architectural families. Convolutional neural networks remain the most extensively studied paradigm in medical image analysis, with multiple surveys documenting their widespread adoption across diagnostic tasks and the evolution of convolutional architectures for classification and segmentation~\cite{review-1, review-5, review-3, review-7, review-8}. These studies primarily emphasize improvements in diagnostic accuracy and architectural design, while deployment-related constraints such as inference latency, memory footprint, and energy consumption receive comparatively limited attention.

The emergence of vision transformers has motivated a parallel body of survey literature, examining their capacity to capture long-range dependencies in medical images and their adaptation to various imaging tasks~\cite{review-4}. Subsequent surveys further analyze transformer variants tailored for medical imaging applications, often acknowledging the substantial computational cost of self-attention mechanisms. However, efficiency considerations are typically discussed within the transformer family itself, and direct comparisons with optimized convolutional architectures or alternative sequence modeling approaches under identical deployment conditions remain uncommon.

More recently, state space models have been proposed as computationally efficient alternatives to transformers. Dedicated surveys review the application of architectures such as Mamba across medical imaging tasks including classification, segmentation, and image restoration, highlighting favorable theoretical scaling properties and architectural design choices that mitigate quadratic complexity~\cite{review-9-mamba, review-10-mamba, review-11-mamba}. Nevertheless, empirical analyses of real-world deployment behavior, including hardware-specific latency, memory usage, and energy efficiency, are still relatively limited.

In addition to general architectural surveys, several focused reviews examine efficiency and compression within specific model families. Surveys on vision transformer compression for edge deployment provide taxonomies of pruning, quantization, and distillation strategies alongside discussions of hardware-aware optimization, while focused analyses of state space models explore design variants with attention to computational complexity and runtime behavior. Although these works offer detailed technical insights within individual paradigms, they do not explicitly frame efficiency trade-offs in relation to clinical deployment constraints such as privacy preservation, latency requirements, and infrastructure heterogeneity.

Collectively, architecture-centric surveys offer in-depth perspectives within individual model families but provide limited guidance on cross-architectural trade-offs that are critical for deployment in clinical environments.

\subsection{Efficiency-Aware and Deployment-Oriented Reviews}

Beyond architecture-centric analyses, a smaller body of literature explicitly addresses efficiency and deployability in medical artificial intelligence. Surveys on edge deep learning and lightweight models discuss compression techniques, hardware constraints, and runtime considerations for point-of-care and mobile devices, emphasizing latency, memory usage, and power consumption~\cite{review-2, review-6, review-12}. However, these evaluations often span broad application domains and rarely establish a unified framework for comparing fundamentally different architectural paradigms under consistent clinical constraints.

In digital pathology, where whole-slide images frequently exceed gigapixel resolution, efficiency challenges are commonly addressed at the pipeline level rather than solely through backbone design. Approaches based on ultra-high-resolution tokenization, feature aggregation, and multiple-instance learning aim to reduce computational and memory requirements prior to or during inference. Recent methods demonstrate how scalable whole-slide image analysis can be achieved without exhaustively processing all patches, enabling global context modeling at substantially lower computational cost in clinically relevant high-resolution settings~\cite{tang2024holohisto,ling2024agent}.

Complementary advances in multiple-instance learning further illustrate deployment-conscious design beyond backbone efficiency. Prototype-based Multiple Instance Learning methods and graph-based coarsening frameworks seek to improve training and inference efficiency while preserving interpretability at the instance or region level, an important consideration in pathology workflows where clinical trust and localized evidence are essential. These approaches operate orthogonally to backbone selection and model compression strategies, underscoring that deployment considerations extend beyond architectural efficiency alone.

Another important aspect of deployment readiness concerns inference toolchains and compiler support. Practical efficiency gains often depend on execution frameworks such as TensorRT, ONNX Runtime, OpenVINO, CoreML, NNAPI, and TVM, which translate trained models into optimized hardware-specific kernels~\cite{NVIDIATensorRT, MicrosoftONNXRuntime, IntelOpenVINO, AppleCoreML, AndroidNNAPI, tvm}. Architectural efficiency does not automatically translate into real-world speedups unless supported by the underlying runtime and compiler ecosystem, and techniques such as structured sparsity, operator fusion, and mixed-precision inference yield measurable benefits only when explicitly supported by the deployment backend. This factor is rarely considered in prior survey literature.

Overall, existing surveys provide valuable insights into individual architectural families, efficiency techniques, and domain-specific deployment strategies, yet the literature remains fragmented with limited synthesis across architectures from a unified deployment-oriented perspective. Rather than revisiting algorithmic details or proposing new models, this review organizes prior work according to how different architectural design choices address practical clinical constraints, including latency, energy consumption, privacy preservation, and hardware accessibility. This perspective motivates the taxonomy of deployment-ready architectures presented in the following section.

%% file: tex/taxonomy.tex
\section{Evolution of Deployment-Ready Architectures: From Inductive Bias to Linear Global Context}
\label{sec:taxonomy}

\begin{figure*}[ht]
\centering
\includegraphics[width=1.0\textwidth]{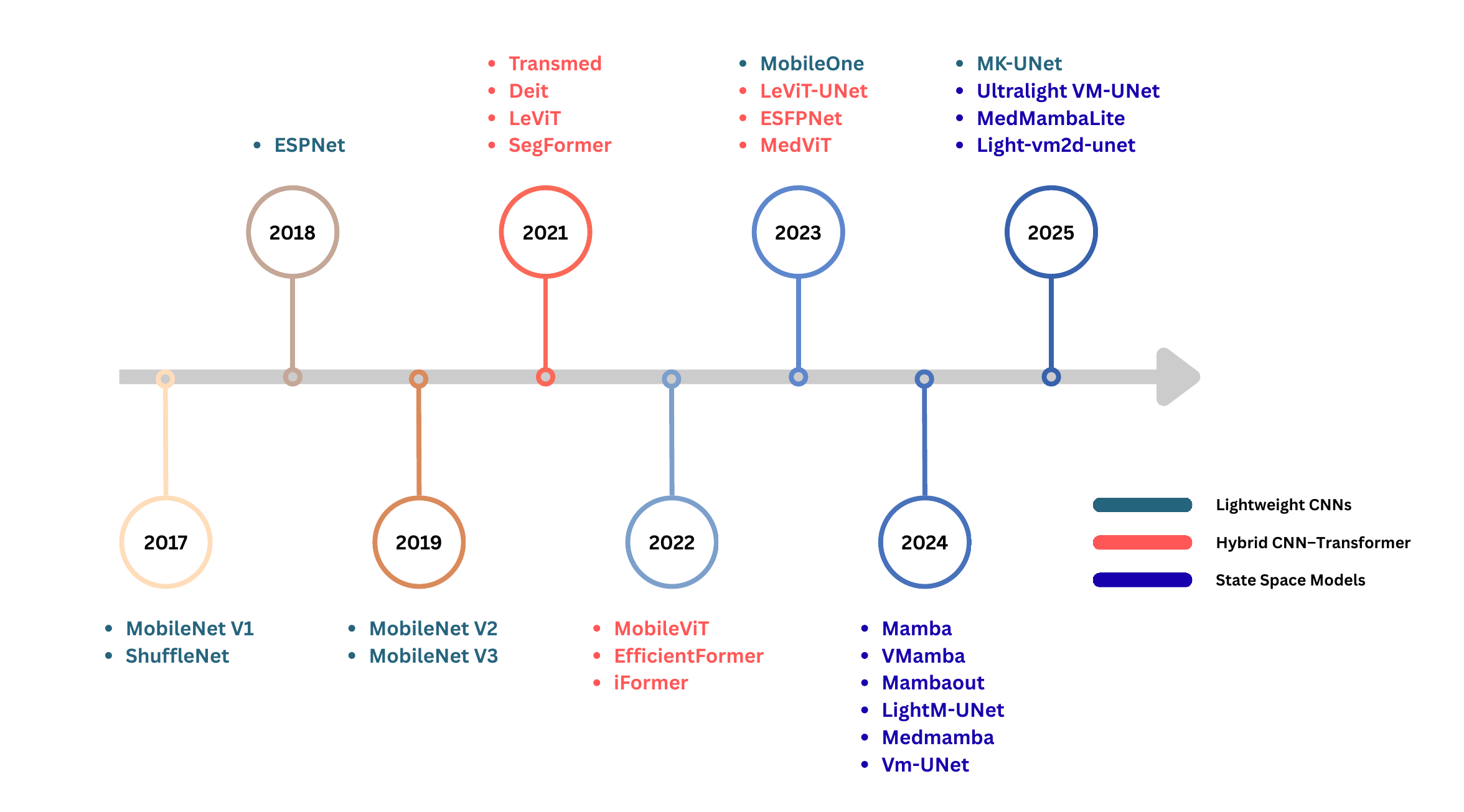}
\caption{Chronological evolution of efficient deep learning architectures in medical imaging, showing the transition from inductive bias-heavy CNNs to global-context Transformers and the recent emergence of linear-complexity State Space Models.}
\label{fig:timeline}
\end{figure*}

The landscape of deployment-ready architectures has evolved through three major technological waves, as illustrated in Figure~\ref{fig:timeline}. This evolution reflects a continuous pursuit of optimal trade-offs between diagnostic fidelity and computational cost, moving from the rigid local features of CNNs to the dynamic global context of Transformers, and most recently to the linear-complexity modeling of State Space Models. 

Before dissecting specific architectural families, however, we must address a pervasive misconception in model evaluation: treating Floating-point Operations (FLOPs) and parameter counts as sole proxies for efficiency. Historically, researchers equated low FLOP counts with deployment readiness, assuming an idealized computing environment where arithmetic intensity dominates performance. Real-world clinical deployment tells a different story, particularly on edge devices like mobile NPUs or embedded GPUs, where latency depends less on calculation volume and more on Memory Access Cost (MAC) and degree of parallelism~\cite{ma2018shufflenetv2practicalguidelines}. This discrepancy creates an efficiency gap: architectures optimized purely for FLOPs often suffer from low hardware utilization through excessive use of fragmented depth-wise convolutions or sparse connections, resulting in inference times that contradict their theoretical lightness. Consequently, the modern definition of deployment readiness has shifted from FLOP-centric minimization to system-level optimization, prioritizing characteristics such as memory contiguity, operator fusion compatibility, and dimension consistency. We adopt this hardware-aware perspective as the analytical lens for reviewing the three architectural paradigms below.

\subsection{CNNs as the Mature Standard for Local Efficiency}

Early successes of deep learning in medical imaging relied on architectures like VGG and ResNet, which prioritized representational capacity over computational feasibility~\cite{simonyan2015deepconvolutionalnetworkslargescale, he2015deepresiduallearningimage}. While effective in retrospective studies utilizing high-end GPUs, these dense, heavy models proved ill-suited for the emerging paradigm of point-of-care and portable diagnostics. To bridge this gap, research focus shifted toward rethinking the fundamental building blocks of CNNs, leading to adoption of Depth-wise Separable Convolutions (pioneered by the MobileNet family~\cite{mobilenets, sandler2019mobilenetv2invertedresidualslinear, howard2019searchingmobilenetv3}) and Channel Shuffling (ShuffleNet~\cite{shufflenet}). By decoupling spatial filtering from channel mixing, these architectures reduced theoretical computational cost by an order of magnitude, establishing the first standard for lightweight medical analysis. However, widespread deployment revealed a critical discrepancy: theoretical efficiency did not guarantee real-world speed, as inference latency on edge devices was often bottlenecked by fragmented memory access rather than arithmetic intensity. Depth-wise convolutions, despite being FLOP-efficient, can suffer from suboptimal bandwidth utilization on certain hardware. Addressing this, recent latency-aware designs like MobileOne~\cite{mobileone} introduce Structural Re-parameterization, a technique that decouples training-time complexity (multi-branch topology for convergence) from inference-time execution (single-path topology for speed), enabling sub-millisecond latency on mobile devices by minimizing MAC and maximizing operator fusion.

Despite these advances, generic lightweight backbones often struggle to capture subtle, irregular structures characteristic of medical images, as fixed, localized receptive fields cannot adequately represent diffuse lesion boundaries or fine vascular networks. This limitation has motivated a wave of task-specific micro-architectures tailored for medical constraints. MK-UNet~\cite{mkunet} and LDMRes-Net~\cite{ldmres} employ multi-kernel depth-wise convolutions to aggregate features at multiple scales simultaneously without the heavy overhead of traditional pyramids, allowing precise boundary delineation in segmentation tasks with extreme parameter efficiency (MK-UNet uses only 0.027M parameters). Similarly, U-Lite~\cite{ulite} leverages Axial Convolutions to expand the receptive field along distinct spatial axes, essential for organ-level context, while remaining $35\times$ smaller than a standard U-Net. In interventional scenarios like endoscopy and ultrasound, where latency supersedes peak accuracy and maintaining high throughput is non-negotiable, ESPNet~\cite{espnet} exemplifies the solution by utilizing Efficient Spatial Pyramids with dilated convolutions. This approach expands the receptive field to capture global context without the resolution loss associated with pooling, enabling segmentation at over 100 FPS on edge GPUs, a critical requirement for closed-loop visual feedback. 

Collectively, these developments establish CNNs as the most mature paradigm for local efficiency, demonstrating through evolution from generic depth-wise separability to hardware-aware re-parameterization and medical-specific micro-designs that diagnostic fidelity can coexist with stringent resource constraints. However, their reliance on localized operators imposes intrinsic limitations in modeling long-range dependencies, setting the stage for the hybrid and attention-based paradigms discussed next.

\subsection{The Hybrid Compromise: Reconciling Local Efficiency with Global Context}

While CNNs have proven robust for local feature extraction, their reliance on finite receptive fields imposes a fundamental limitation in modeling long-range dependencies, a constraint particularly consequential in medical imaging where diagnostic reasoning often depends on spatial relationships between distant anatomical structures (consider identifying global symmetry in brain MRI or contextualizing diffuse pathologies like interstitial lung disease). Vision Transformers (ViTs) theoretically resolve this by capturing global interactions via self-attention~\cite{vit}, yet their quadratic complexity ($O(N^2)$) and lack of inductive bias render them computationally prohibitive for resource-constrained edge devices. To bridge this gap, Hybrid Architectures have emerged as a pragmatic compromise, strategically embedding attention mechanisms within convolutional pipelines to amortize the cost of global context modeling across efficient local representations. The evolution of hybrid designs reflects a critical shift from simple architectural integration to hardware-aware co-design: early approaches such as MobileViT~\cite{mobilevit} treated Transformers as drop-in modules, unfolding feature maps into non-overlapping patches to capture global context before refolding them for spatial processing, but recent deployment analyses reveal that these frequent reshaping operations create significant compiler friction on mobile NPUs, as switching between 4D-tensor and 3D-sequence representations often bottlenecks actual inference latency.

Addressing this, next-generation hybrids like EfficientFormerV2~\cite{efficientformer2} prioritize dimension-consistent design by retaining a 4D convolutional structure for the majority of the network and restricting computationally expensive attention mechanisms to the final stages, minimizing reshaping overhead and allowing effective leverage of mobile compilers (CoreML, SNPE) to achieve latencies comparable to MobileNet while retaining the global reasoning of Transformers. Similarly, iFormer~\cite{iformer} tackles the unique frequency characteristics of medical images by employing an Inception-like mixer that captures high-frequency information (local textures, edges) and low-frequency information (global shape, structure) in parallel paths, optimizing capture of fine lesion boundaries without the latency penalty of dense self-attention. Beyond classification, hybrid architectures have proven particularly adept at dense prediction tasks essential for clinical workflows: in volumetric segmentation where fixed input sizes are rarely guaranteed, SegFormer~\cite{xie2021segformersimpleefficientdesign} eliminates rigid positional encodings in favor of a hierarchical Mix-Transformer encoder, allowing robust generalization across varying image resolutions and slice thicknesses, while architectures like LeViT-UNet~\cite{xu2023levit} further optimize speed by aggressively integrating convolutions and Batch Normalization into the Transformer encoder, reducing memory access overhead to enable segmentation speeds of up to 114 FPS and validating that attention mechanisms, when carefully constrained, are compatible with latency-critical interventional applications such as real-time endoscopy analysis~\cite{esfpnet}.

Finally, the hybrid paradigm offers distinct advantages in data efficiency and multi-modal reasoning, addressing the scarcity of annotated medical data: through token-based distillation strategies like those in DeiT~\cite{touvron2021trainingdataefficientimagetransformers}, hybrids inherit the strong inductive biases of CNN teachers, enabling faster convergence on small datasets where pure Transformers often struggle, while frameworks like TransMed~\cite{dai2021transmed} leverage the attention mechanism to model cross-modal dependencies (such as the correlation between T1-weighted and T2-weighted MRI sequences), outperforming single-modality baselines that lack the capacity to fuse heterogeneous feature spaces. 

In summary, hybrid architectures successfully balance the trade-off between local efficiency and global context, though their reliance on attention mechanisms, even when optimized, still incurs non-trivial memory access costs for high-resolution or long-sequence inputs, motivating exploration of linear-complexity paradigms such as State Space Models discussed in the next section.

\subsection{The Linear Revolution: State Space Models and Mamba}

Although hybrid CNN--Transformer architectures substantially reduce the practical cost of global context modeling, they do not fundamentally resolve the quadratic scaling of self-attention: as image resolution increases, attention-based models still incur rapidly growing memory and latency overheads, limiting their applicability to gigapixel histopathology slides, long cine MRI sequences, or high-resolution 3D volumetric scans. This bottleneck has motivated a search for architectures that can preserve global receptive fields while eliminating quadratic complexity altogether, culminating in the decisive shift to State Space Models (SSMs) for deep learning, particularly the Mamba architecture~\cite{mamba}. Originally developed for long-range sequence modeling, Mamba offers a compelling proposition for medical imaging: global information propagation with strictly linear computational complexity ($O(N)$), conceptually bridging the gap between recurrent models and Transformers by replacing explicit attention with selective state updates that enable long-context reasoning without pairwise token interactions. This property positions SSMs as a fundamentally different and potentially more scalable solution to the deployment challenges faced by attention-based models, though adapting Mamba to medical images is non-trivial, as the original formulation operates on 1D sequences whereas medical data are inherently 2D or 3D and exhibit strong spatial locality.

Early efforts addressed this mismatch through structured scanning strategies that linearize images while preserving spatial coherence: VMamba introduced the Cross-Scan Module (CSM), which traverses feature maps along four directions (horizontal, vertical, and their reverses), allowing the SSM to aggregate contextual information across spatial axes without explicit attention~\cite{vmamba}, a design that has emerged as a commonly adopted design choice for vision-oriented SSMs, particularly in dense prediction tasks. Despite their promise, SSMs are not universally superior to attention-based models, as critical analysis (MambaOut) shows that the advantages of SSMs are highly task-dependent~\cite{mambaout}: for dense prediction problems such as segmentation, where accurate delineation of long-range boundaries is essential, SSMs consistently outperform purely convolutional baselines, whereas for image-level classification tasks with limited spatial complexity, gated CNNs or lightweight hybrids may achieve comparable performance with lower architectural complexity. This distinction highlights an important theme: efficiency gains from SSMs stem not only from novelty, but also from alignment between the architectural inductive bias and task structure. The most striking impact of Mamba-based models appears in resource-constrained segmentation scenarios, where LightM-UNet integrates Mamba blocks into a U-Net-style architecture to achieve state-of-the-art segmentation accuracy with approximately 1M parameters~\cite{lightmunet}, while Ultralight VM-UNet pushes efficiency to an extreme by employing a Parallel Vision Mamba (PVM) layer to deliver competitive performance with only 0.049M parameters and 0.06 GFLOPs ~\cite{wu2025ultralight}, representing an extreme point in the current efficiency landscape for deployable medical segmentation models and highlighting results that would be unattainable under quadratic attention scaling.

Beyond parameter efficiency, recent work explores biologically inspired scanning strategies to improve feature retention: Light-VM2D-UNet introduces Wavefront Scanning, which propagates information in a manner analogous to physical signal diffusion through tissue~\cite{light-vm2d-unet}, preserving anatomical continuity more effectively compared to raster or cross-scan traversal and suggesting that scanning order is not merely a computational artifact but a meaningful design choice in medical SSMs. For classification tasks, hybridization remains beneficial, as MedMamba interleaves convolutional layers with SSM blocks to combine the strong inductive bias of CNNs with the long-range modeling capability of Mamba~\cite{medmamba}, while this design philosophy extends naturally to deployment-oriented optimization through MedMambaLite, which further distills and re-parameterizes the architecture for edge accelerators to achieve a reported 63\% improvement in energy efficiency on platforms such as NVIDIA Jetson Orin Nano compared to its non-optimized teacher model~\cite{medmambalite}. This result emphasizes that the value of SSMs lies not only in speed or parameter count, but also in enabling sustainable, battery-powered medical AI systems aligned with Green AI objectives. 

In summary, Mamba-based SSM architectures represent a qualitative departure from attention-centric design: by replacing quadratic interactions with linear state propagation, SSMs enable global context modeling at a scale previously incompatible with clinical deployment constraints, and while they are not a universal replacement for CNNs or hybrid models, their effectiveness in high-resolution and resource-limited settings positions them as a foundational building block for the next generation of efficient medical imaging systems.

\subsection{Comparative Analysis and Hardware Benchmarking}

The evolution from CNNs to Mamba represents a continuous search for the optimal Pareto frontier between diagnostic fidelity and computational cost. To provide a holistic view of this landscape, we synthesize the architectural trade-offs of each paradigm in Table~\ref{tab:unified_backbone_taxonomy}, a qualitative taxonomy that highlights how successive generations have shifted from rigid inductive biases (CNNs) to selective attention (Hybrids) and finally to linear global context (SSMs), each addressing specific clinical limitations. However, architectural elegance does not guarantee deployment efficiency, as discrepancies between theoretical complexity (FLOPs) and physical latency often arise from hardware-specific constraints, memory access patterns, and compiler maturity. 

To ground these architectural trade-offs in deployment reality, Table~\ref{tab:hardware_benchmark} organizes representative models into two categories: those with \textit{verified mobile latency measurements} and those demonstrating \textit{parameter efficiency without platform-specific benchmarks}. This division reflects a critical methodological constraint in current literature. While parameter counts are architecture-intrinsic properties, latency is a hardware-compiler co-design outcome that requires explicit measurement under controlled conditions. The scarcity of standardized edge-device benchmarks, particularly for recent SSM-based architectures, prevents comprehensive cross-paradigm latency comparison and motivates our focus on models with reproducible deployment evidence.

\begin{table*}[htbp]
\centering
\small
\caption{Unified taxonomy of deployment-oriented backbone architectures in medical imaging.}
\label{tab:unified_backbone_taxonomy}
\renewcommand{\arraystretch}{1.4}
\begin{tabularx}{\textwidth}{
    >{\RaggedRight\hsize=0.7\hsize}X 
    >{\RaggedRight\hsize=0.75\hsize}X 
    >{\RaggedRight\hsize=1.0\hsize}X 
    >{\RaggedRight\hsize=1.3\hsize}X 
    >{\RaggedRight\hsize=1.25\hsize}X
}
\toprule
\textbf{Paradigm} & \textbf{Representative Models} & \textbf{Core Design Principle} & \textbf{Key Deployment Strengths} & \textbf{Intrinsic Limitations \& Clinical Trade-offs} \\
\midrule
\textbf{Lightweight CNNs} & 
MobileNet, ShuffleNet, MobileOne, ESPNet, MK-UNet, U-Lite & 
Strong inductive bias via locality and weight sharing; efficiency through depth-wise, grouped, and re-parameterized convolutions & 
\begin{itemize}[nosep, leftmargin=*, after=\strut]
    \item Lowest latency on edge devices
    \item Hardware-friendly memory access patterns
    \item Mature toolchain support (CPU, GPU, NPU)
    \item Proven real-time performance in ultrasound and endoscopy
\end{itemize} & 
\begin{itemize}[nosep, leftmargin=*, after=\strut]
    \item Limited long-range dependency modeling
    \item Global context requires architectural workarounds (dilation, pyramids)
    \item Performance degrades in multi-organ or volumetric reasoning tasks
\end{itemize} \\
\midrule
\textbf{Hybrid CNN--Transformer} & 
MobileViT, LeViT, SegFormer, LeViT-UNet, ESFPNet, MedViT, EfficientFormer, iFormer & 
Selective global context modeling via constrained self-attention embedded within convolutional pipelines & 
\begin{itemize}[nosep, leftmargin=*, after=\strut]
    \item Balanced local efficiency and global awareness
    \item Better performance under limited annotations
    \item Effective for multi-scale and multi-organ segmentation
    \item Achievable real-time inference when attention is restricted
\end{itemize} & 
\begin{itemize}[nosep, leftmargin=*, after=\strut]
    \item Attention mechanisms remain memory and latency intensive
    \item Performance sensitive to resolution and tokenization strategies
    \item Harder to optimize across heterogeneous hardware platforms
\end{itemize} \\
\midrule
\textbf{State Space Models} & 
VMamba, LightM-UNet, Ultralight VM-UNet, MedMambaLite & 
Linear-complexity global context propagation via state updates instead of pairwise attention & 
\begin{itemize}[nosep, leftmargin=*, after=\strut]
    \item True global receptive field with $O(N)$ complexity
    \item Exceptional parameter and energy efficiency
    \item Scales effectively to high-resolution and long sequences
    \item Strong fit for battery-powered and Green AI deployment scenarios
\end{itemize} & 
\begin{itemize}[nosep, leftmargin=*, after=\strut]
    \item Weaker spatial inductive bias compared to CNNs
    \item Performance sensitive to scan order and architectural design choices
    \item Less mature software and hardware ecosystem
\end{itemize} \\
\bottomrule
\end{tabularx}
\end{table*}
\begin{table*}[htbp]
\centering
\caption{Quantitative benchmarking of representative deployment-ready architectures. Metrics are reported from original publications with verified experimental conditions. Latency measurements are hardware- and compiler-specific; direct cross-platform comparisons should account for toolchain differences.}
\label{tab:hardware_benchmark}
\scriptsize
\setlength{\tabcolsep}{4pt}
\begin{tabular}{llllccc}
\toprule
\textbf{Model} & \textbf{Task} & \textbf{Paradigm} & \textbf{Target Hardware} & \textbf{Input Res.} & \textbf{Latency} & \textbf{Params (M)} \\
\midrule
\multicolumn{7}{c}{\textit{Verified Mobile NPU Benchmarks}} \\
\midrule
MobileOne-S0~\cite{mobileone} & Classif. & CNN & iPhone 12 NPU & $224 \times 224$ & \textbf{0.79 ms} & 2.1 \\
EfficientFormerV2-S0~\cite{efficientformer2} & Classif. & Hybrid & iPhone 12 NPU & $224 \times 224$ & 0.9 ms & 3.5 \\
MobileViT-XS~\cite{mobilevit} & Classif. & Hybrid & iPhone 12 NPU & $256 \times 256$ & 7.3 ms & 2.3 \\
\midrule
\multicolumn{7}{c}{\textit{Representative Lightweight Architectures}} \\
\midrule
ESPNet~\cite{espnet} & Segment. & CNN & -- & -- & 100+ FPS\textsuperscript{\dag} & 0.36 \\
LightM-UNet~\cite{lightmunet} & Segment. & SSM & -- & -- & -- & 1.0 \\
Ultralight VM-UNet~\cite{wu2025ultralight} & Segment. & SSM & -- & -- & -- & \textbf{0.049} \\
LeViT-UNet-128~\cite{xu2023levit} & Segment. & Hybrid & -- & -- & 114 FPS\textsuperscript{\dag} & 15.9 \\
VM-UNet-B~\cite{vm-unet} & Segment. & SSM & -- & -- & -- & 27.4 \\
MedMamba-B~\cite{medmamba} & Classif. & SSM & -- & -- & -- & $\sim$24 \\
\bottomrule
\end{tabular}
\begin{minipage}{\textwidth}
\vspace{1mm}
\small
\textbf{Note:} Latency measurements for mobile NPU models use CoreML compilation on iPhone 12 as reported in original publications. \textsuperscript{\dag}FPS metrics reported without specific hardware context in source papers. Missing values (--) indicate absence of quantitative measurements in original publications or inability to verify experimental conditions. Input resolutions are reported where explicitly stated in source papers; missing resolutions reflect variability in evaluation protocols across studies. Parameter counts reflect architectural properties independent of deployment platform.
\end{minipage}
\end{table*}

Among architectures with verified iPhone 12 NPU latency, pure CNNs like MobileOne-S0 establish the performance ceiling at sub-millisecond inference (0.79 ms), while dimension-consistent hybrids like EfficientFormerV2-S0 demonstrate that carefully designed Transformers can approach this efficiency frontier with only 14\% overhead (0.9 ms), contrasting sharply with earlier designs like MobileViT-XS that incur 8$\times$ slowdown (7.3 ms) due to tensor reshaping operations fragmenting computational graphs. Meanwhile, models like Ultralight VM-UNet achieve remarkable parameter compression (0.049M), yet lack corresponding latency validation on edge devices, reflecting the infrastructural reality that SSM-based selective state-space operators often rely on custom CUDA kernels without native support in mobile inference frameworks such as CoreML, TensorFlow Lite, or Qualcomm SNPE. This sparse latency landscape underscores broader methodological fragmentation in efficient deep learning research: reported metrics vary across batch sizes, warmup protocols, numerical precision, and backend engines, while energy consumption measurements range from unreliable software profiling to rarely documented instrumented power rails, precluding unified efficiency ranking without standardized protocols analogous to MLPerf~\cite{reddi2020mlperf}. Consequently, for immediate clinical deployment on battery-powered devices, only architectures with verified mobile latency represent evidence-based choices, while SSM-based and ultra-lightweight models offer compelling parameter-accuracy trade-offs for server-side inference, though practitioners should benchmark candidates on target hardware rather than relying on reported metrics from potentially incomparable experimental setups. The future viability of SSM architectures at the point of care depends on two critical developments: the maturation of selective scan kernels in mobile inference engines and the establishment of community-wide benchmarking standards that enable reproducible latency claims.

%% file: tex/optimization.tex
\section{Balancing Efficiency and Fidelity: Model Compression in Medical AI}
\label{sec:techniques}

The deployment gap between state-of-the-art deep learning models and real-world clinical environments is often not architectural, but computational in nature. Even when efficient backbones such as lightweight CNNs, hybrid Transformers, or Mamba are employed, the resulting models frequently exceed the memory, latency, or energy budgets of PoC devices. Consequently, model compression must be viewed not merely as a post hoc optimization, but as a foundational requirement for the democratization of medical AI~\cite{choudhary2020comprehensive,dantas2024comprehensive,li2023model}.

To address these constraints, contemporary literature converges on four orthogonal but complementary paradigms: network pruning, quantization, knowledge distillation, and low-rank factorization. Table \ref{tab:compression_summary} provides a high-level comparison of these strategies, summarizing their core mechanisms and the critical trade-offs between computational efficiency and diagnostic safety. In the following sections, we examine the specific implementation and clinical implications of each approach.

\subsection{Network Pruning}

Network pruning reduces computational complexity by eliminating parameters, channels, or tokens that contribute minimally to the inference process. In medical imaging, pruning is particularly appealing due to the inherent sparsity of clinically relevant information. Large portions of scans often contain background, air, or healthy tissue.

A representative example is APFormer, proposed by Lin et al., which introduces adaptive token pruning into Transformer-based segmentation~\cite{lin2023lighter}. Unlike uniform attention mechanisms that process all patches equally, APFormer dynamically discards redundant tokens corresponding to non-informative regions during inference. This region-aware pruning achieves substantial GFLOP reductions while preserving segmentation accuracy, illustrating how task structure can guide safe compression.

However, pruning introduces unique risks in medical contexts. Holste et al.\ demonstrated that on long-tailed datasets, where rare pathologies are underrepresented, pruning can disproportionately impair the detection of minority classes, even when aggregate metrics such as AUC remain unchanged~\cite{holste2023does}. These ``pruning-identified exemplars'' (PIEs) expose a critical limitation: aggressive sparsification may silently erode sensitivity to rare but clinically significant findings. Consequently, pruning strategies must be evaluated beyond global performance metrics, with explicit attention to safety-critical edge cases.

\subsection{Quantization}

Quantization reduces model size and accelerates inference by lowering numerical precision, typically converting FP32 representations to INT8, FP16, or even ternary formats. For edge devices equipped with NPUs, quantization is often the most direct pathway to real-time performance.

Early studies demonstrated that post-training INT8 quantization enables up to 97\% latency reduction for cancer diagnosis models deployed on portable devices, transforming offline pipelines into real-time clinical tools~\cite{garifulla2021case}. Subsequent work identified 8-bit precision as a practical compromise, achieving 4--8$\times$ parameter reduction while retaining over 80\% diagnostic accuracy~\cite{ahamad2022quantized}.


Nevertheless, quantization is not without trade-offs. Reducing numerical precision may attenuate fine-grained signal representations, which can be critical in medical images where diagnostically relevant patterns are often subtle. Quantized models may also exhibit increased sensitivity to data distribution shifts, particularly when deployment conditions differ from those used for calibration. Additionally, very low-bit implementations often rely on hardware-specific support, which can limit cross-platform generalizability and complicate large-scale clinical deployment.

\subsection{Knowledge Distillation}

Knowledge Distillation (KD) compresses models by transferring knowledge from a large, well-trained teacher model to a compact student model. Unlike pruning or quantization, KD does not remove capacity directly; instead, it reshapes the learning objective so that lightweight models inherit clinically meaningful decision boundaries.

Alabbasy et al.\ demonstrated that distilled students deployed via TensorFlow Lite can retain diagnostic accuracy within 1--2\% of heavyweight teachers, validating KD as a deployment-ready strategy~\cite{alabbasy2023compressing}. More recent work tailors student architectures to specific clinical domains. ConvMixer-based students have been designed for pulmonary disease detection, while global context from Vision Transformer teachers has been distilled into convolutional U-Net students, effectively embedding long-range anatomical reasoning into efficient CNNs~\cite{asham2025lightweight,song2024medical}. 

Advanced distillation schemes address specific compression pathologies. Rank-Sensitive Knowledge Distillation (RSKD) preserves attention hierarchy during compression, maintaining global contextual understanding despite reduced model capacity~\cite{liang2024rskd}. Similarly, KD-enabled denoising frameworks reconstruct diagnostically reliable images on resource-constrained edge devices~\cite{muksimova2023enhancing}.

However, KD's suitability for medical imaging introduces critical dependencies and failure modes that warrant careful consideration. The student model inherits not only the teacher's capabilities but also its systematic biases and calibration errors, a particularly concerning issue in medical contexts where silent error propagation can compromise diagnostic safety without triggering obvious performance degradation during validation. The distillation process itself introduces substantial methodological complexity through hyperparameter sensitivity across temperature scaling, loss weighting, and intermediate supervision layers, making reproducible cross-task deployment difficult and limiting the generalizability of distillation protocols. Furthermore, architectural heterogeneity between teachers and students creates a fundamental bottleneck in knowledge transfer. When student capacity or inductive biases differ significantly from the teacher, certain representational structures cannot be faithfully compressed, potentially losing clinically important feature interactions that lack explicit supervision signals in standard distillation objectives. This structural mismatch becomes especially problematic in medical imaging, where subtle spatial relationships and rare pathological patterns may reside precisely in those hard-to-transfer representational spaces that conventional distillation losses fail to capture.

\subsection{Low-Rank Factorization}

Low-rank factorization approximates large convolutional or attention matrices using products of smaller matrices, reducing both parameter count and computational complexity. Beyond compression, this technique acts as an implicit regularizer, which is advantageous for medical datasets that are often limited in size.

Low-rank constraints have been applied to 3D convolutional kernels in volumetric brain tumor segmentation, substantially reducing model size while maintaining competitive Dice scores~\cite{ashtari2020low}. By explicitly modeling redundancy in high-dimensional medical features, low-rank factorization achieves compression without relying on post hoc sparsification, offering a principled alternative for volumetric imaging tasks.

However, the benefits of low-rank factorization are contingent on the extent to which redundancy is truly present in the underlying feature representations. When pathological patterns are spatially localized or highly heterogeneous, aggressive rank constraints may oversimplify the feature space, leading to a loss of discriminative capacity. In practice, selecting an appropriate rank often requires empirical tuning and task-specific insight, which can limit the method’s robustness and transferability across diverse medical imaging modalities.

\begin{table*}[t]
\centering
\small
\caption{Comparison of model compression strategies for deployable medical AI.}
\label{tab:compression_summary}
\renewcommand{\arraystretch}{1.4}
\begin{tabularx}{\textwidth}{
    >{\RaggedRight\hsize=0.65\hsize}X 
    >{\RaggedRight\hsize=0.85\hsize}X 
    >{\RaggedRight\hsize=1.25\hsize}X 
    >{\RaggedRight\hsize=1.25\hsize}X
}
\toprule
\textbf{Method} & \textbf{Core Mechanism} & \textbf{Key Advantages} & \textbf{Limitations and Risks} \\
\midrule
Network Pruning &
Eliminates structurally or functionally redundant parameters, channels, or tokens &
\begin{itemize}[nosep, leftmargin=*, after=\strut]
    \item Significant reduction in FLOPs and memory footprint
    \item Naturally aligned with spatial sparsity in medical images
    \item Enables dynamic, region-aware inference
\end{itemize} &
\begin{itemize}[nosep, leftmargin=*, after=\strut]
    \item Risk of degrading sensitivity to rare or long-tailed pathologies
    \item Performance loss may not be reflected in aggregate metrics
    \item Requires careful safety-aware evaluation
\end{itemize} \\
\midrule
Quantization &
Reduces numerical precision of weights and activations &
\begin{itemize}[nosep, leftmargin=*, after=\strut]
    \item Direct acceleration on edge NPUs
    \item Substantial reductions in model size and energy consumption
    \item Often compatible with post-training deployment
\end{itemize} &
\begin{itemize}[nosep, leftmargin=*, after=\strut]
    \item Attenuation of fine-grained signal representations
    \item Increased sensitivity to distributional shift
    \item Hardware and platform-dependent implementations
\end{itemize} \\
\midrule
Knowledge Distillation &
Transfers representational and decision knowledge from teacher to student models &
\begin{itemize}[nosep, leftmargin=*, after=\strut]
    \item Preserves clinically meaningful decision boundaries
    \item Flexible across architectures and tasks
    \item Supports safety-aware compression
\end{itemize} &
\begin{itemize}[nosep, leftmargin=*, after=\strut]
    \item Strong dependence on teacher quality and calibration
    \item Increased training complexity and hyperparameter sensitivity
    \item Partial loss of representational knowledge under architectural mismatch
\end{itemize} \\
\midrule
Low-Rank Factorization &
Approximates high-dimensional operators using low-rank matrix products &
\begin{itemize}[nosep, leftmargin=*, after=\strut]
    \item Principled reduction of algebraic redundancy
    \item Acts as an implicit regularizer for small datasets
    \item Well-suited to volumetric and high-dimensional data
\end{itemize} &
\begin{itemize}[nosep, leftmargin=*, after=\strut]
    \item Performance depends on true redundancy in features
    \item Risk of oversimplification for heterogeneous pathologies
    \item Requires task-specific rank selection
\end{itemize} \\
\bottomrule
\end{tabularx}
\end{table*}

%% file: tex/discussion.tex
\section{Discussion}

\subsection{Navigating the Architectural Landscape: Mamba, Hybrids, and the 3D Bottleneck}

Rather than exhaustively cataloging architectural variants, this discussion focuses on the practical implications of recent design choices for deployment-oriented medical imaging systems. A central dilemma for practitioners in the current era is selecting an appropriate backbone amid the growing divergence between attention-based and state-space architectures. Increasingly, this decision is shaped not only by accuracy, but by the alignment between a model’s inductive bias and the spatial structure of the target modality.

For high-resolution 2D modalities such as whole-slide pathology or chest X-ray, the linear complexity of Mamba-based models offers a compelling advantage under memory-constrained settings. However, empirical studies suggest that the choice of scan strategy plays a critical role in translating this theoretical efficiency into practice. While wavefront-style scans are conceptually appealing and biologically motivated, emerging evidence across recent works indicates that cross-scan strategies tend to provide a more stable balance between long-range context aggregation and memory access efficiency on contemporary hardware. These observations highlight scan order as a task-dependent and modality-dependent design choice rather than a universally optimal solution.

In 3D volumetric imaging tasks such as CT and MRI, the trade-offs become more nuanced. Here, deployment feasibility is often governed by memory bandwidth rather than raw compute. Under memory-limited regimes (e.g., GPUs with less than 8GB VRAM), pure or predominantly SSM-based designs, such as lightweight Mamba-UNet variants, appear better suited to processing larger volumetric patches without incurring the quadratic memory growth associated with full attention mechanisms. Conversely, for diagnostic tasks that hinge on subtle local texture cues, such as detecting early ground-glass opacities, hybrid Transformer architectures may remain a more conservative choice, leveraging the well-established representational strengths of convolutional stems combined with selective attention, provided hardware resources permit.

A promising emerging direction is the ``asymmetric hybrid'' paradigm, in which efficient 2D SSM encoders are used for slice-wise feature extraction, followed by lightweight 3D aggregation to enforce volumetric consistency. Such designs respect the anisotropic nature of many medical volumes while offering a pragmatic compromise between accuracy, throughput, and memory consumption.

\subsection{The ``Compiler Gap'': Bridging Theory and Real-World Latency}

Despite rapid architectural innovation, a persistent gap remains between theoretical efficiency and realized deployment performance. Our review highlights that commonly reported metrics such as FLOPs and parameter counts often fail to predict end-to-end latency or energy consumption in clinical systems. This discrepancy stems largely from the uneven maturity of software toolchains supporting different operator families.

Convolutional networks currently enjoy a decisive deployment advantage, benefiting from decades of optimization in mature ecosystems such as NVIDIA TensorRT, OpenVINO, and CoreML, where depthwise and grouped convolutions are routinely fused and scheduled for efficient execution. In contrast, many Mamba-based and SS2D-based models rely on custom kernels that lack native support in widely used inference engines, including ONNX Runtime and TFLite Micro. As a result, architectures that are mathematically efficient may exhibit suboptimal performance on heterogeneous or non-NVIDIA edge platforms, such as mobile DSPs or low-power CPUs.

This “compiler gap” has direct clinical implications: unpredictable latency and energy behavior can undermine real-time decision support and disrupt workflow integration at the point of care. In the near term, highly optimized CNN-based models therefore remain the most reliable choice for large-scale deployment across diverse hardware fleets. Closing this gap for SSM-based architectures will require a coordinated co-design effort spanning model architecture, operator standardization, and compiler development, potentially leveraging emerging frameworks such as TVM or Triton to bridge research prototypes and production systems.

\subsection{Safety-Critical Optimization: Quantization and Calibration}

As medical AI systems are increasingly pushed toward edge deployment, aggressive optimization techniques such as quantization become difficult to avoid. In healthcare settings, however, optimization must be framed not merely as an engineering problem, but as a safety-critical design decision. Reducing numerical precision from FP32 to INT8 can substantially improve latency and energy efficiency, yet it may also degrade uncertainty calibration or disproportionately affect sensitivity to rare but clinically significant findings.

Importantly, a quantized model may retain strong aggregate performance metrics, such as global Dice or AUC, while becoming subtly overconfident in incorrect predictions or less responsive to long-tail pathologies. To address this risk, we argue that the evaluation of deployment-ready models should routinely extend beyond accuracy to include safety-oriented indicators such as Expected Calibration Error (ECE), Negative Log Likelihood (NLL), and rare-class sensitivity, particularly in post-quantization settings.

Across the reviewed literature, Quantization-Aware Training consistently emerges as more reliable than Post-Training Quantization for medical imaging tasks, as it allows models to adapt their decision boundaries to reduced-precision representations. In addition, mixed-precision strategies, preserving higher precision in the first and last layers while aggressively quantizing intermediate blocks, offer a pragmatic compromise that maintains diagnostic fidelity while delivering meaningful efficiency gains. Reporting such design choices and their calibration impact should be considered best practice rather than optional.

\subsection{The Imperative of Explainability in Compact Models}

As models become smaller and more aggressively optimized, the role of explainability shifts from a supplementary feature to a core component of validation. Lightweight architectures produced through pruning, distillation, or quantization risk achieving efficiency by exploiting dataset-specific shortcuts rather than learning clinically meaningful representations. In this context, explainability mechanisms play a crucial role in verifying that compression has not eroded the causal basis of model predictions.

For edge-deployed systems operating with limited or no human-in-the-loop oversight, explainable AI serves as an important safety valve. Post hoc techniques such as saliency mapping or attention visualization can support auditing and error analysis, helping clinicians assess whether automated decisions align with anatomical or pathological expectations. More broadly, these considerations motivate growing interest in intrinsic explainability, designing efficient architectures whose internal representations are interpretable by construction, rather than relying exclusively on post hoc analysis to justify opaque models.

\subsection{Synergy Between Efficiency, Privacy, and Equity}

The implications of architectural efficiency extend beyond individual device performance to influence the broader ecosystem of medical AI deployment. There is a natural synergy between lightweight models and privacy-preserving paradigms such as federated or split learning. In collaborative training settings, communication bandwidth often represents the dominant bottleneck; here, compact architectures, low-rank adaptation, and quantized updates can substantially reduce synchronization costs, enabling participation by institutions with limited infrastructure.

Efficiency also plays a central role in promoting equity. By enabling advanced diagnostic models to operate on battery-powered or low-cost hardware, efficient AI systems can help extend imaging-based decision support to resource-constrained and rural environments. From this perspective, the shift toward compact and energy-efficient architectures is not merely a technical optimization, but a foundational step toward more sustainable, accessible, and inclusive digital health systems.

\subsection{Toward a Deployment-First Evaluation Standard for Medical Imaging Models}

As efficiency-driven architectures increasingly target edge and point-of-care deployment, the field faces a critical tension between the demand for quantitative evidence and the absence of standardized evaluation practices. While recent studies routinely report latency, memory footprint, or energy metrics, these numbers often emerge from heterogeneous and poorly specified conditions. Variations in hardware platforms, runtime engines, numerical precision, batch sizes, and warm-up protocols can induce order-of-magnitude performance differences. Without explicit disclosure, such variability risks being misattributed to architectural superiority rather than recognized as an artifact of evaluation setup.

This challenge is particularly acute in medical imaging, where deployment constraints are operational rather than abstract. A model labeled ``efficient'' under GPU-centric benchmarking may fail real-time requirements on CPU-only workstations, mobile NPUs, or embedded accelerators. Conversely, architectures optimized for specific runtimes may appear disproportionately strong when evaluated with favorable toolchains. Absent a common reporting baseline, cross-study comparison becomes unreliable, leaving clinicians and system designers without context to assess whether reported efficiency gains transfer to their deployment environments.

Commonly reported proxies (FLOPs, parameter count, or theoretical GOPS/J) provide incomplete views of deployability. FLOPs ignore memory access patterns and operator fusion; GOPS/J is rarely measured under controlled power instrumentation and conflates architectural properties with hardware-specific characteristics. When reported in isolation, these metrics provide partial and sometimes misleading characterizations of real-world performance.

Optimization techniques further complicate evaluation. Quantization, pruning, and operator re-parameterization can dramatically reduce latency and energy consumption while potentially degrading calibration, rare-class sensitivity, or numerical stability, clinically meaningful properties seldom reported alongside efficiency gains. Two models with identical latency profiles may differ substantially in diagnostic reliability, a distinction invisible under conventional benchmarking.

Rather than proposing another centralized benchmark, we advocate for a deployment-first reporting standard: minimal but explicit disclosure of conditions under which efficiency claims are obtained. This standard does not mandate uniform hardware or datasets but makes efficiency claims interpretable by anchoring them to clearly defined hardware–software stacks, inference protocols, and measurement procedures.

Table~\ref{tab:deployment_checklist} presents a proposed checklist for reporting efficiency-oriented results. This framework is intentionally lightweight, not a prescriptive benchmark protocol requiring multi-device evaluation, but a set of contextual disclosures enabling readers to situate metrics within their operational scope. By making assumptions explicit (task definition, runtime engines, precision, measurement procedures), authors reduce ambiguity while preserving flexibility.
Importantly, this approach acknowledges uncertainty as intrinsic to deployment evaluation. Reporting median and tail latency (e.g., p90), warm-up behavior, and cross-run variance recognizes that edge performance is non-deterministic, particularly on shared or thermally constrained devices. Similarly, documenting calibration strategies and post-quantization reliability metrics reframes efficiency as a safety-critical property, not merely a computational achievement.

\begin{table*}[htbp]
\centering
\caption{Proposed Deployment-First Reporting Checklist for Efficiency-Oriented Medical Imaging Models. Emphasizes contextual disclosure rather than prescriptive benchmarking, enabling interpretable comparison across heterogeneous edge environments.}
\label{tab:deployment_checklist}
\scriptsize
\setlength{\tabcolsep}{4pt}
\begin{tabular}{p{3.2cm} p{12.8cm}}
\toprule
\textbf{Category} & \textbf{Recommended Disclosure Items} \\
\midrule
Task & Clinical task (classification, segmentation, triage), modality (CT, MRI, US, WSI), input resolution, preprocessing pipeline, dataset scale, class imbalance \\
Hardware Platform & Device model (CPU/GPU/NPU/FPGA), memory capacity, power mode (sustained vs burst), thermal constraints, system load conditions \\
Software Stack & Runtime engine (TensorRT, ONNX Runtime, OpenVINO, CoreML, TFLite), version numbers, backend configuration, compiler optimizations \\
Numerical Precision & Inference precision (FP32/FP16/INT8), quantization method (PTQ vs QAT), calibration dataset size, per-tensor vs per-channel quantization \\
Inference Protocol & Batch size (recommended = 1 for edge), warm-up runs, timed runs, synchronization method, I/O handling scope \\
Latency Reporting & Median and tail metrics (p50, p90), variance across runs, notes on throttling or performance drift \\
Memory Footprint & Model size on disk, peak runtime memory, additional runtime/compiler buffers \\
Energy \& Power & Measurement method (on-device sensors or external), idle power subtraction, joules per inference, variance \\
Reliability \& Safety & Post-optimization calibration (ECE, NLL), rare-class sensitivity, robustness to noise/resolution shifts, failure mode analysis \\
\bottomrule
\end{tabular}
\end{table*}

To demonstrate this framework's utility, we provide a reference implementation benchmarking two representative efficient architectures, MobileOne-S0 (pure CNN) and EfficientFormerV2-S0 (Hybrid), under standardized deployment assumptions. Table~\ref{tab:benchmark_config} documents the exact evaluation context, while Table~\ref{tab:benchmark_results} reports the measured performance metrics across both CPU and GPU substrates.

\begin{table}[htbp]
\centering
\caption{Benchmark Configuration and Measurement Environment. Explicit disclosure of hardware specifications, software versions, and measurement protocols ensures reproducibility.}
\label{tab:benchmark_config}
\small
\renewcommand{\arraystretch}{1.2} 
\begin{tabular}{llp{6cm}}
\toprule
\textbf{Category} & \textbf{Parameter} & \textbf{Specification} \\
\midrule
\textbf{Target Task} & Task Type & Image Classification (ImageNet-1k) \\
& Input Resolution & $224 \times 224 \times 3$ \\
& Batch Size & 1 (Simulating real-time edge deployment) \\
\midrule
\textbf{Hardware} & CPU & Intel Core i7-12700K (12 Cores, 3.6--5.0 GHz) \\
& System Memory & 32GB DDR4-3200 \\
& GPU Accelerator & NVIDIA GeForce RTX 3070 Ti (8GB GDDR6X) \\
\midrule
\textbf{Software Stack} & OS & Linux x86\_64 (Ubuntu 22.04 LTS) \\
& Python Environment & Python 3.12 \\
& DL Framework & PyTorch 2.x (CUDA 12.x backend for GPU) \\
\midrule
\textbf{Instrumentation} & Latency Timing & \texttt{torch.cuda.Event} (GPU) / \texttt{time.perf\_counter} (CPU) \\
& Energy Profiling & \texttt{perf stat -e power/energy-pkg/} (CPU Package) \\
& & \texttt{nvidia-smi dmon} (GPU Power Draw) \\
\midrule
\textbf{Protocol} & Precision & FP32 (Single Precision) \\
& Sampling & 50 Warm-up runs + 200 Measured runs \\
\bottomrule
\end{tabular}
\end{table}

\begin{table}[htbp]
\centering
\caption{Deployment-Oriented Efficiency Metrics. The table contrasts performance on commodity CPU vs. accelerated GPU hardware. Speedup factors in parentheses are calculated relative to the CPU baseline.}
\label{tab:benchmark_results}
\small
\setlength{\tabcolsep}{3.5pt} 
\begin{tabular}{lcccccc}
\toprule
\textbf{Model} & \textbf{Params} & \textbf{Median Latency} & \textbf{P90 Latency} & \textbf{Throughput} & \textbf{Energy} & \textbf{Peak RSS} \\
& \textbf{(M)} & \textbf{(ms)} & \textbf{(ms)} & \textbf{(img/s)} & \textbf{(J/inf)} & \textbf{(MB)} \\
\midrule
\multicolumn{7}{l}{\textit{\textbf{Scenario A: CPU Inference} (Intel i7-12700K)}} \\
MobileOne-S0 & 2.08 & 5.18 & 5.36 & 193 & 0.94 & 832 \\
EfficientFormerV2-S0 & 3.60 & 11.70 & 11.94 & 85 & 1.64 & 901 \\
\midrule
\multicolumn{7}{l}{\textit{\textbf{Scenario B: GPU Inference} (RTX 3070 Ti)}} \\
MobileOne-S0 & 2.08 & \textbf{1.19} ($4.3\times$) & 1.23 & \textbf{842} & \textbf{0.60} & 1156 \\
EfficientFormerV2-S0 & 3.60 & 4.34 ($2.7\times$) & 4.44 & 230 & 0.67 & 1267 \\
\bottomrule
\end{tabular}
\end{table}

These results, grounded in the declared context, reveal critical deployment dynamics. First, despite higher thermal design power, GPU inference validates the "race-to-idle" principle by consuming less net energy ($0.60$~J) than the CPU ($0.94$~J) due to its faster execution, thereby minimizing the total power integral. Second, architectural affinity plays a measurable role: the pure CNN (MobileOne) achieves a significantly higher GPU acceleration factor ($4.3\times$) than the hybrid EfficientFormerV2 ($2.7\times$), suggesting that dense convolutions leverage massive parallelism more effectively than attention mechanisms at a batch size of one.

Crucially, these performance rankings are strictly bound by the specific hardware--software stack. Altering runtime engines (e.g., TensorRT), quantization schemes, or target processors could fundamentally shift the observed ordering. Ultimately, this standardized disclosure prioritizes reproducibility over absolute rankings, enabling readers to distinguish fundamental architectural properties from environment-specific optimizations.

%% file: tex/conclusion.tex
\section{Conclusion} \label{sec:conclusion}

This review has traced the evolution of deployment-ready architectures for medical imaging, from convolutional models with strong local inductive bias to hybrid CNN-Transformer designs and, most recently, to linear-complexity state space models. Across this progression, a central theme emerges: practical clinical impact depends not only on representational power, but on how effectively models align with real-world constraints on latency, memory, energy, and hardware heterogeneity.

Lightweight CNNs remain the most mature and reliable solution for low-latency edge deployment, while hybrid architectures offer a pragmatic balance between efficiency and global contextual reasoning. State space models such as Mamba introduce a fundamentally new design paradigm, enabling scalable global context modeling under strict resource budgets and opening new opportunities for high-resolution and energy-efficient medical AI.

Looking forward, the translation of these architectures into routine clinical use will require a shift toward deployment-first evaluation, task-specific and context-specific optimization, and greater emphasis on interpretability and robustness. By integrating architectural innovation with system-level efficiency and clinical trust, future lightweight models can move beyond benchmark performance and support sustainable, equitable, and trustworthy medical AI deployment.